%% file: main.tex
\documentclass[10pt,twocolumn,letterpaper]{article}
\usepackage[pagenumbers]{cvpr} 

\input{preamble}

\definecolor{cvprblue}{rgb}{0.21,0.49,0.74}
\usepackage[pagebackref,breaklinks,colorlinks,citecolor=cvprblue]{hyperref}

\title{EventNeuS: 3D Mesh Reconstruction from a Single Event Camera} 

\author{
Shreyas Sachan$^{1,2}$ \;
Viktor Rudnev$^{1,2}$ \;
Mohamed Elgharib$^{2}$ \;
Christian Theobalt$^{2}$ \;
Vladislav Golyanik$^{2}$\\[0.5em]
$^{1}$Saarland University, SIC \quad
$^{2}$MPI for Informatics, SIC 
}

\begin{document}
\input{sec/0_teaser}
\input{sec/0_abstract}    
\input{sec/1_intro}
\input{sec/2_related}
\input{sec/3_background}
\input{sec/4_method}
\input{sec/5_experiments}
\input{sec/6_conclusions}

{
    \small
    \bibliographystyle{ieeenat_fullname}
    \bibliography{main}
}

\input{sec/X_supp}
\end{document}

%% file: preamble.tex
\usepackage{multirow}
\usepackage[export]{adjustbox}
\usepackage[dvipsnames]{xcolor}
\renewcommand{\paragraph}[1]{\noindent\textbf{#1.}}

\usepackage{times}
\usepackage{epsfig}
\usepackage{graphicx}
\usepackage{amsmath}
\usepackage{amssymb}
\usepackage{siunitx}
\usepackage{comment}

%% file: sec/0_teaser.tex
\twocolumn[{%
\renewcommand\twocolumn[1][]{#1}%
\maketitle
\centering
\vspace{-1.5em}
\includegraphics[width=1\linewidth]{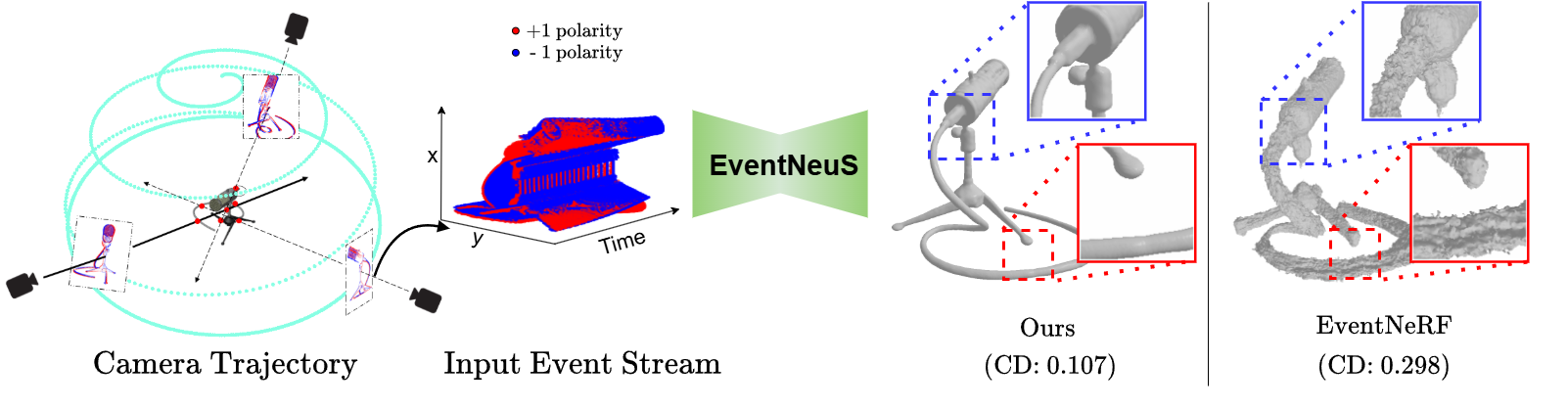}
\vspace{-20pt}
\captionof{figure}{Left: A moving monocular event camera capturing the asynchronous per-pixel brightness changes, with red (positive) and blue (negative) polarities representing pixel intensity changes over time. Centre: The captured event stream is processed by the proposed \textbf{EventNeuS} to reconstruct a detailed 3D mesh. Right: Compared to the previous state of the art ~\cite{eventnerf}, our method produces a more accurate reconstruction with a lower Chamfer Distance (CD: 0.107 vs 0.298), highlighting its effectiveness in the event-based 3D reconstruction. 
\vspace{2em}
}
\label{fig:teaser} 
}]

%% file: sec/0_abstract.tex
\begin{abstract} 
Event cameras offer a considerable alternative to RGB cameras in many scenarios. While there are recent works on event-based novel-view synthesis, dense 3D mesh reconstruction remains scarcely explored and existing event-based techniques are severely limited in their 3D reconstruction accuracy. To address this limitation, we present EventNeuS, a self-supervised neural model for learning 3D representations from monocular colour event streams. Our approach, for the first time, combines 3D signed distance function and density field learning with event-based supervision. Furthermore, we introduce spherical harmonics encodings into our model for enhanced handling of view-dependent effects. EventNeuS outperforms existing approaches by a significant margin, achieving 34\% lower Chamfer distance and 31\% lower mean absolute error on average compared to the best previous method\footnote{Project page: \url{https://4dqv.mpi-inf.mpg.de/EventNeuS/}}. 
\end{abstract}

%% file: sec/1_intro.tex
\section{Introduction}
\label{sec:intro}

Most existing methods for 3D reconstruction rely on RGB inputs and techniques such as Structure-from-Motion \cite{wang2023vggsfm, he2024detector, lindenberger2021pixel} and Multi-View Stereo \cite{sayed2022simplerecon, shi2023raymvsnet++, zhang2023regionaware} for depth estimation and surface reconstruction. 
While the current state of the art achieves high 3D reconstruction accuracy, it often struggles with rapid motion or varying lighting conditions, leading to motion blur and inaccuracies in 3D reconstruction. 
To address these limitations, event-based methods have been recently explored, leveraging the high temporal resolution and dynamic range of event cameras \cite{Chakravarthi2024}. 
While event cameras offer advantages in high-speed settings, existing event-based 3D reconstruction techniques result in sparse representations and cannot reconstruct dense scene surface details. 
Moreover, many of these methods depend on explicit feature matching ~\cite{zhou2018semi, messikommer2023data}, which is challenging considering the sparse and asynchronous nature of event data. 
It can further limit the reconstruction accuracy   
or require additional synchronous RGB recordings, 
potentially diminishing the utility of event observations and the multimodal setup overall. 
Hence, there remains a need for approaches that can fully exploit the unique capabilities of event cameras to achieve dense and accurate 3D surface reconstruction, particularly methods that operate exclusively with event data without the compromises often required in multimodal setups. 
This work proposes EventNeuS, a new method for detailed surface reconstruction solely from monocular event streams. 
We capitalise on the insights from the recent literature on event-based novel-view synthesis \cite{eventnerf} and surface extraction from neural radiance fields \cite{neus, unisurf}. 
EventNeuS leverages a neural implicit Signed Distance Function (SDF) \cite{deepsdf} to implicitly model scene geometry using per-colour-channel changes captured by a single event camera. 
It accumulates events over time in event frames, effectively capturing the dynamic changes occurring within the scene, as illustrated in \cref{fig:teaser}. 
The event frames supervise a coordinate-based neural network that learns an SDF of the observed scene. 
The SDF maps any 3D point of the reconstructed volume to its distance from the nearest surface, enabling high-fidelity surface extraction upon convergence. 
We introduce several innovations compared to previous techniques. 
First---to improve the modelling of view-dependent effects---we replace conventional positional encoding for camera view directions with Spherical Harmonics (SH) encodings, applying them in event-based 3D reconstruction for the first time. 
Second, our novel self-supervised architecture enables joint learning of a neural radiance field (NeRF) \cite{nerf} and an SDF of a scene relying solely on camera-pose-informed event-based supervision. 
We design a loss function that aligns temporal differences in the rendered RGB images with the accumulated event data, effectively capturing the per-pixel changes over time and enabling neural network weight optimisation. 
Additionally, we incorporate regularisation terms to ensure smoothness and consistency across the reconstructed surfaces. 
In summary, our technical contributions are as follows: 
\begin{itemize}[leftmargin=15pt]
 \item[1)] EventNeuS, the first method to extract high-quality 3D meshes \textbf{solely} from monocular event streams, with no RGB frames or explicit feature matching, enabling dense surface reconstruction (Sec.~\ref{sec:approach}); 
 \item[2)] Novel integration of \textbf{spherical harmonics encoding, hierarchical importance sampling, and frequency-annealing technique} for capturing view-dependent effects and learning fine geometric details from sparse event data (Sec.~\ref{sec:volume rendering}); 
 \item[3)] A new \textbf{spiral-trajectory synthetic dataset for 3D mesh reconstruction} from event streams (Sec.~\ref{sec:experiments}). 
\end{itemize}

EventNeuS outperforms multiple competing methods by at least ${\approx}34\%$ on the Chamfer distance metric. 
The numerical evaluation also translates into substantial and consistent qualitative improvements in all tested scenarios. 
%
%

%% file: sec/2_related.tex
\section{Related Works}
\label{sec:related}
\paragraph{Event-based 3D Reconstruction} 
Most event-based 3D reconstruction approaches reconstruct sparse point clouds in the context of simultaneous localisation and mapping ~\cite{Kim2016, RebecqEVO, hidalgo2022event, zhou2021event}. 
Some approaches apply event cameras for photometric stereo capturing detailed surface normals, i.e.,~partial aspects of 3D object geometry \cite{eventps}. 
Only a few methods for dense event-based 3D reconstruction of rigid ~\cite{evac3d} and non-rigid \cite{eventCap,eventhpe,eventhands,Millerdurai_3DV2024, Millerdurai_EventEgo3D_2024,eventegoplusplus,nakabayashi2025ev4dgs} scenes exist. 
EvAC3D ~\cite{evac3d} reconstructs object shapes from events, assuming those are generated by object silhouettes. 
It continuously carves a visual hull from apparent event contours, and, like any silhouette method, often misses fine geometric details. 
Next, Event-ID ~\cite{eventid} employs SDF for implicit geometry reconstruction from multi-modal inputs, i.e.,~event stream and blurry RGB images. 
Our EventNeuS approach, in contrast, achieves detailed 3D reconstruction from events alone, without RGB supervision or prior shape templates. 
We adopt recent advances in neural fields for novel-view rendering and 3D reconstruction from multi-view 3D inputs \cite{neus, Yariv2021, unisurf}. 
While RGB-based methods achieve impressive 3D fidelity, they rely on standard cameras and struggle with fast motion or low-light conditions. 
Our EventNeuS adapts for the first time the principles of implicit surface learning to the setting with event stream supervision, enabling 3D reconstruction under challenging conditions. 
\paragraph{Event-based Novel-View Synthesis} 
EventNeRF ~\cite{eventnerf} pioneered NeRF learning from event streams. 
It provoked several follow-ups \cite{evnerf, enerf, cannici2024mitigating, low2023robust, low2024deblur, li2024benerf}, similarly enabling novel view rendering in the RGB space under low-light conditions from fast-moving event cameras. 
While all these approaches produce accurate novel views, they are not designed for accurate 3D surface reconstruction. 
While these approaches produce plausible novel views, they do not explicitly enforce constraints facilitating surface learning; the scene geometry remains implicit in the density field, leading to oversmoothed or missed fine details. 
PAEv3D ~\cite{paev3d} augments event-based NeRF with motion and geometry priors to improve reconstruction stability, yet the emphasis is still on rendering quality over exact geometry. 
While 3D shapes can be extracted from EventNeRF and its extensions using Marching Cubes \cite{LorensenCline1987}, it is well known that the extracted geometry can be highly rugged and insufficiently accurate for many applications. 
In contrast, our method models surfaces via an SDF in addition to volumetric scene properties, which results in the recovery of more accurate 3D shape details compared to the previous state of the art.  

%% file: sec/3_background.tex
\section{Background}
\label{sec:background}
\subsection{Event Accumulation for Learning}
\label{subsec:accumulation}

Event-based cameras record asynchronous per-pixel intensity changes.
Each pixel independently tracks a reference value for the logarithmic brightness \( L(t) = \log_{\gamma} I(t) \), where \( I(t) \in \mathbb{R}^{W \times H} \) is the absolute intensity image at time \( t \), \( W \) and \( H \) are the image dimensions, and \( \gamma \) is a gamma correction factor. 
When the logarithmic brightness change at a pixel reaches a predefined threshold \( C \), the camera triggers an event \( E_i = (x_i, y_i, t_i, p_i) \), where \( (x_i, y_i) \) denotes pixel coordinates, \( t_i \) is the timestamp, and \( p_i \in \{-1, +1\} \) encodes the polarity. 
Each event corresponds to a fixed brightness change:
\begin{equation}
L_{x_i, y_i}(t_i) - L_{x_i, y_i}(t_{i-1}) = p_i C,
\label{eq:event_generation}
\end{equation}
where \( t_{i-1} \) is the timestamp of the previous event at the same pixel \( (x_i, y_i) \), with \( p_i = +1 \) for the brightness increase and \( p_i = -1 \) for the decrease. 
Existing event accumulation strategies segment events into discrete time windows bounded by \( t_i = i / N_{\text{windows}} \) for \( i \in \{1, \dots, N_{\text{windows}}\} \) \cite{eventnerf}. 
Fixed windows face a trade-off: Constant short windows preserve local details but struggle to propagate high-level shape details, whereas constant long windows better capture object shape but can oversmooth high-frequency appearance details. 
An adaptive strategy can be used with the window start \( t_0 \) uniformly sampled from \( \mathcal{U}[t - L_{\text{max}}, t) \), with \( L_{\text{max}} \) as the maximum window length. 
This stochastic sampling enables the joint modelling of high-frequency colour details and low-frequency lighting effects without sacrificing temporal resolution. 

\subsection{Wang et al.'s NeuS~\cite{neus}}
\label{sec:NeuS}

Recent advances in neural implicit representations have enabled high-fidelity 3D surface reconstruction from multi-view images. Among these, NeuS \cite{neus} introduces a framework that combines Signed Distance Functions (SDFs) with volume rendering to recover surfaces as the zero-level set:
\begin{equation}
S = \left\{ \mathbf{x} \in \mathbb{R}^3 \mid f_{\text{sdf}}(\mathbf{x}) = 0 \right\},
\end{equation}
where \( f_{\text{sdf}} \colon \mathbb{R}^3 \to \mathbb{R} \) is a neural network predicting signed distances.
Traditional volume rendering applies density-based ray marching to SDFs, but the naive policy would introduce bias because the density peak often fails to align with the true surface location. 
NeuS addresses this by redefining the weighting function to ensure the rendered colour correlates directly with the SDF’s zero-crossing. 
Given a camera ray \( \mathbf{r}(t) = \mathbf{o} + t\mathbf{d} \) with origin \( \mathbf{o} \) and direction \( \mathbf{d} \), NeuS computes the density \( \sigma(t) \) as:
\[
\alpha_i = \max\left( \frac{\Phi_s(f_{\text{sdf}}(t_i)) - \Phi_s(f_{\text{sdf}}(t_{i+1}))}{\Phi_s(f_{\text{sdf}}(t_i))}, 0 \right),
\]
where \( \Phi_s(x) = (1 + e^{-sx})^{-1} \) is a learnable sigmoid function scaled by parameter \( s \), which controls the sharpness of the SDF-to-density transition. The accumulated transmittance \( T(t) \) and pixel colour \( \hat{C} \) are then computed as 
\begin{align}
T(t) &= \exp\left( -\int_{0}^{t} \alpha(\tau) \, d\tau \right)\,\text{and}\\
\hat{C} &= \int_{0}^{\infty} T(t) \alpha(t) c(\mathbf{r}(t), \mathbf{d}) \, dt,
\end{align}
where \( c(\mathbf{r}(t), \mathbf{d}) \) is the view-dependent radiance predicted by the second neural network.

While NeuS’s weighting scheme is theoretically unbiased, real-world factors (e.g.,~sparse camera views, occlusions, or SDF estimation errors) can degrade reconstruction accuracy. 
While this method is likely to output high-quality reconstruction under dense and noiseless inputs, it requires additional regularisation for noisy or sparse data, as follow-ups demonstrated~\cite{li2023neuralangelo, zhang2022critical}. 
Its probabilistic formulation ensures that surface estimates remain close to the true geometry even when biases arise from imperfect observations. 

%% file: sec/4_method.tex
\section{The EventNeuS Approach} 
\label{sec:approach} 

\begin{figure*}
    \centering
    \includegraphics[width=1.0\textwidth,trim={0cm 0cm 0cm 0cm},clip]{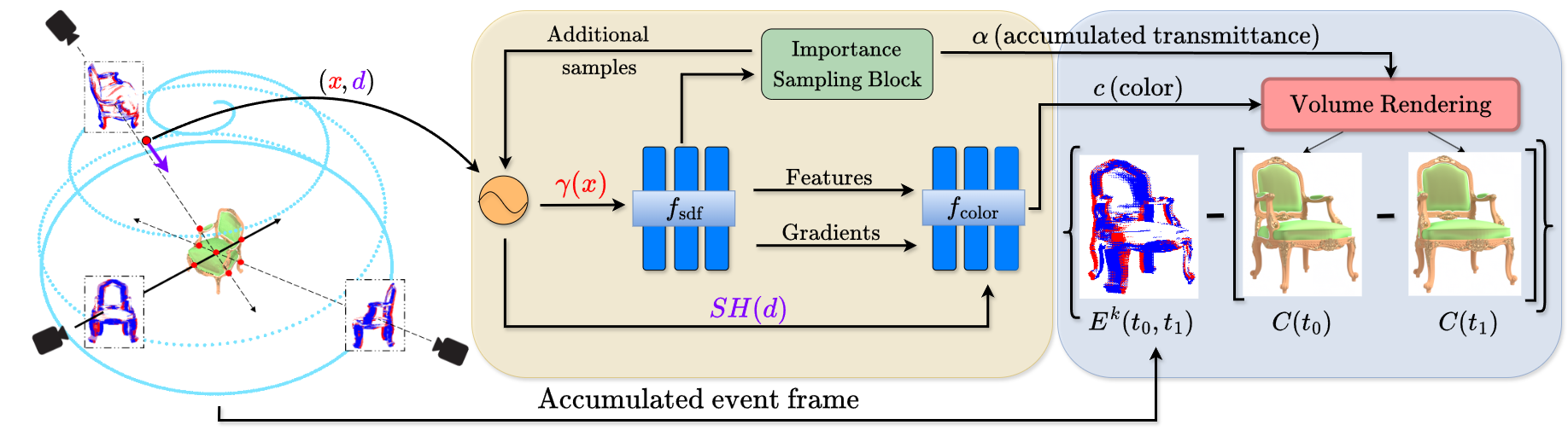}
    \vspace{-25pt}
    \caption {Overview of \textbf{EventNeuS}.
      We start with the trajectory that captures the object from multiple viewpoints (e.g.,~Seiffert's spherical spiral). 
      We accumulate all events within a time window $[t_0, t_1]$ to form an event frame $E^k(t_0, t_1)$ (\cref{subsec:accumulation}). 
      For each event frame, we randomly choose a mini-batch of pixels and sample points along the corresponding rays.
      After applying positional encoding (\cref{sec:freq-anneal}) to these 3D coordinates, we feed them into the $f_{\text{sdf}}$ network, which outputs a Signed Distance Function (SDF) and its gradients.
      We further refine our sampling near surfaces via importance sampling (\cref{sec:imp-sampling}).
      Next, we combine the resulting SDF features and gradients with view directions encoded via spherical harmonics $SH(d)$ in the $f_{\text{colour}}$ network to predict colour (\cref{sec:sh_encoding}).
      Finally, we convert the SDF to a density field using \cref{sec:sdf2density}, integrate it along each ray to obtain the \emph{accumulated transmittance} $\alpha$, and use the colour predictions to get the \emph{rendered colour} $c$ through volumetric rendering.
      We render two views (at the start and at the end of the event window) and take their difference, applying an Mean Squared Error (MSE) loss against the ground-truth accumulated event frame (~\cref{sec:self-supervision}).
      This difference enforces consistency between the rendered scene changes and the actual events recorded during $[t_0, t_1]$.
    }
    \label{fig:fig1}
\end{figure*}

We propose a method to reconstruct a 3D object mesh from a single event stream, eliminating the need for RGB data during training; see Fig.~\ref{fig:fig1} for an overview. 
At the core of our approach is a neural implicit Signed Distance Function (SDF) $\varphi(\mathbf{x})$, which maps any 3D point $\mathbf{x} \in \mathbb{R}^3$ to its signed distance from the nearest surface. 
We adapt NeuS \cite{neus} to handle the sparse and temporally irregular nature of event data by directly linking brightness\footnote{we refer to brightness and not colour channel changes because the colours are ordered according to the Bayer filter on the pixel grid} changes to the geometric and photometric properties of the reconstructed 3D object. 
Following EventNeRF \cite{eventnerf}, we apply a Bayer filter to our predictions when computing the loss, such that only one colour channel per pixel is supervised, matching the camera model. 
To model view-dependent effects efficiently, we replace conventional positional encoding for camera view directions with Spherical Harmonics (SH)—marking the first use of SH in event-based 3D reconstruction to handle view-dependent effects. 
SH compactly represents low-frequency view variations, reducing computational overhead while improving surface detail. 
Following the accumulation strategy in EventNeRF (\cref{subsec:accumulation}), we convert asynchronous events into temporal windows by aggregating pixel-wise polarities.
Each window encodes brightness changes as sparse frames. We assume known camera intrinsics and extrinsics for each window to ensure correct 3D-to-2D projection during training. 
Unlike RGB-based methods, we introduce a loss function (\cref{sec:self-supervision}) that aligns temporal differences in rendered images with accumulated event data. 
Specifically, we minimise the mismatch between (1) brightness changes predicted by our radiance field over a time interval and (2) the observed event polarities, leveraging the event camera’s inherent sensitivity to temporal gradients.
Next, to synthesise images from the learned SDF and radiance field, we employ volume rendering with hierarchical sampling and frequency annealing (\cref{sec:volume rendering}). We progressively increase positional encoding frequency bands during training (\cref{sec:freq-anneal}), avoiding early overfitting to noise while recovering fine details. 
We compute opacity $\alpha(t)$ from the SDF gradient using a logistic distribution (\cref{sec:sdf2density}). The final pixel colour $\hat{C}$ integrates radiance values along rays, weighted by transmittance and opacity, to reconstruct surfaces consistent with event-based brightness constraints.
By retaining NeuS’s theoretical guarantees while operating solely on event data, our approach enables robust 3D reconstruction, as validated in \cref{sec:experiments}.

\subsection{Event-Driven Neural Implicit Surface Learning} 
One of the core technical challenges of our setting is adapting NeuS for the SDF $\varphi(\mathbf{x})$ estimation from event streams. 
We design a loss function (\cref{sec:self-supervision}) that aligns temporal differences in rendered images with accumulated event data. Let $\hat{\mathbf{C}}(t_0)$ and $\hat{\mathbf{C}}(t_1)$ denote images synthesised via volume rendering at consecutive timestamps.
Minimising the discrepancy between $\hat{\mathbf{C}}(t_1) - \hat{\mathbf{C}}(t_0)$ and the observed event polarities enforces geometric consistency with the event camera’s brightness change measurements (see \cref{eq:event_generation}).
We model the scene geometry using an MLP approximating $\varphi(\mathbf{x})$, mapping 3D coordinates $\mathbf{x}$ to signed distances. 
To capture high-frequency geometric details, we apply positional encoding to 
$\mathbf{x}$. 
For appearance modelling, a separate MLP predicts view-dependent colours $c(\mathbf{x}, \mathbf{d})$ using SH encoding for view directions $\mathbf{d}$. 
This radiance field is conditioned on geometric features from the SDF network and surface normals $\mathbf{n} = \nabla \varphi(\mathbf{x})$. 
SH encoding efficiently represents low-frequency view variations, reducing computational overhead and stabilising training by avoiding overfitting to sparse event noise. 
We compute opacity $\alpha(t)$ from the SDF gradient via a logistic distribution. 
Colours along rays are integrated using transmittance-weighted radiance values, yielding surfaces consistent with event-based brightness constraints. 
While NeuS assumes photometric stability across RGB frames, we directly link temporal rendering differences to event-driven brightness changes, enabling self-supervised training. 

\subsection{Self-Supervised Training with Event Data}
\label{sec:self-supervision}

Our self-supervised training objective aligns the temporal differences in the rendered images with the accumulated event data.
For each camera view \( k \), we synthesise RGB images \(\hat{\mathbf{C}}^k(t_0)\) and \(\hat{\mathbf{C}}^k(t_1)\) at consecutive timestamps \( t_0 \) and \( t_1 \).
These predictions are aligned with the accumulated event frame \( E^k(t_0, t_1) \in \mathbb{R}^{H \times W} \), which encodes brightness changes over the interval \([t_0, t_1]\).
Following EventNeRF \cite{eventnerf}, we compute the event loss \( \mathcal{L}_{\text{event}} \) in logarithmic space to match the photometric response of event cameras:
{\small
\begin{align}
&L_{\text{event}} =\\ 
&\operatorname{MSE} \Big( F \odot E^k(t_0, t_1),\ 
 F \odot \Big( \log\left( \hat{\mathbf{C}}^k(t_1) \right) \nonumber 
 - \log\left( \hat{\mathbf{C}}^k(t_0) \right) \Big) \Big), \label{eq}
\end{align}
}
\hspace{-6pt}where ``$\operatorname{MSE}(\cdot)$'' is the mean squared error operator, ``$\odot$'' denotes elementwise matrix multiplication, \( \hat{\mathbf{C}}^k(t) \in \mathbb{R}^{H \times W \times 3} \) is the model's predicted RGB image at time \( t \), and \( F \) is a Bayer filter mask applied to simulate the colour sensor array. The logarithmic transform linearises intensity changes, ensuring compatibility with event camera measurements.
To regularise the SDF, we apply the Eikonal loss, which enforces unit gradient norms to maintain valid SDF properties, mitigates surface artefacts and ensures smooth geometry:
\begin{equation} L_{\text{eik}} = \frac{1}{N} \sum_{i=1}^{N} \left( \left| \nabla \varphi(x_i) \right|_2 - 1 \right)^2, \end{equation}
where \( N \) is the number of sampled 3D points \( \mathbf{x}_i \). 

The total training loss combines both components:
\begin{equation}
L_{\text{total}} = L_{\text{event}} + \lambda_{\text{eik}} L_{\text{eik}},
\end{equation}
where \( \lambda_{\text{eik}} = 0.1 \) balances the regularisation strength, determined empirically across all scenes.

\subsection{Volume Rendering with Hierarchical Sampling}
\label{sec:volume rendering}

To synthesise images from the learned SDF and radiance fields, we employ a volume rendering strategy adapted from NeuS (\cref{sec:NeuS}). 
Our approach extends NeuS's unbiased density transformation by introducing hierarchical importance sampling, focusing on regions with richer details along each ray. 
Additionally, we use a frequency annealing strategy during training to progressively improve the stability and accuracy of our model’s geometry predictions. 

\subsubsection{Importance Sampling}
\label{sec:imp-sampling}

We extend NeuS's unbiased density estimation with a hierarchical sampling strategy. 
Unlike NeuS and VolSDF \cite{Yariv2021}, which use density-based PDFs derived from their respective weight functions, we concentrate samples in regions with high event activity or large SDF gradients, exploiting the event camera's inherent sensitivity to edges and texture boundaries. 
Initial coarse samples are drawn uniformly along each ray to estimate surface locations. 
Unlike NeRF’s separate coarse and fine networks, we use a single network for iterative refinement, concentrating additional samples in regions with high event activity or large SDF gradients. 
This adaptive sampling matches NeuS's efficiency (\cref{sec:NeuS}) while focusing computational resources on regions critical for event-based reconstruction. 

\subsubsection{Frequency Annealing}
\label{sec:freq-anneal}

To avoid overfitting and instability due to high-frequency noise in the early stages of training, we incorporate a frequency annealing strategy, gradually introducing high-frequency components as training progresses. 
This approach mitigates issues such as noisy normal maps and inaccurate geometry predictions by controlling the complexity of the frequency bands over time. 
The annealing coefficient \(\beta_k(n)\) for the \(k\)-th frequency band at iteration \(n\) reads: 
\begin{equation}
\beta_k(n) = \frac{1}{2} \left(1 - \cos\left( \pi \cdot \text{clamp}\left( \alpha(n) - k + N_{\text{fmin}}, \ 0, \ 1 \right) \right) \right),
\end{equation}
where $\alpha(n) = (N_{f_{\text{max}}} - N_{f_{\text{min}}}) \cdot \frac{n}{N_{\text{anneal}}}$. Here, \(N_{\text{fmin}}\) (initial bands) and \(N_{f_{\text{max}}}\) (final bands) control the progressive introduction of frequencies, while \(N_{\text{anneal}} = 3 \cdot 10^4\) determines the annealing duration. 
This phased approach first stabilises low-frequency geometry before recovering fine details, reducing artefacts such as noisy surface normals. 

\subsubsection{Spherical Harmonics for View-Dependence}  \label{sec:sh_encoding} 

Event cameras measure temporal brightness changes influenced by view-dependent specular reflections.
These event streams exhibit high sensitivity to sensor noise, which traditional view encoding methods often misinterpret as geometric detail through overfitting. 
We propose spherical harmonics (SH) as the principled encoding basis for view directions, contrasting with standard positional encoding (PE).
While PE operates effectively within unit-box coordinates, it becomes suboptimal for directional quantities: view directions inherently reside on the unit sphere.
We address this by encoding view directions into a 16-dimensional SH basis. 
This SH representation is concatenated with surface normals $\nabla\varphi(\mathbf{x})$ and geometric features from the SDF network $\varphi(\mathbf{x})$.
By integrating SH-encoded directions into our radiance network $f_{\text{colour}}$, we resolve the ambiguity between transient artefacts and true geometric detail, enabling robust reconstruction of surfaces from sparse event streams. 
Implementation details can be found in App.~\ref{ssec:implementation_details}. 
\subsubsection{Opacity and Colour Integration}
\label{sec:sdf2density}

We next convert SDF values to densities \(\sigma(t)\) using a logistic distribution peaking near the zero-level set. 
Opacities \(\alpha_i\) along each ray are computed as:
\begin{equation}
\alpha_i = \max\left( \frac{\Phi_s(f_{\text{sdf}}(t_i)) - \Phi_s(f_{\text{sdf}}(t_{i+1}))}{\Phi_s(f_{\text{sdf}}(t_i))}, 0 \right),
\end{equation}
where \(\Phi_s\) is the sigmoid function from \cref{sec:NeuS}. The final pixel colour integrates radiance values \(c_i\) weighted by transmittance \(T_i\) and opacity \(\alpha_i\):

\[
\hat{I}_t = \sum_{i=1}^N T_i \alpha_i c_i, \quad\text{and}\quad T_i = \prod_{j=1}^{i-1} (1 - \alpha_j).
\]

This formulation ensures that surfaces reconstructed from event data maintain photometric consistency with observed brightness changes.

\subsection{Architecture and Implementation Details}

Our architecture follows the NeuS \cite{neus} design, using two specialised MLPs that jointly model geometry and appearance, with modifications for event-based supervision. 
Once EventNeuS is trained, we query $f_\text{sdf}$ and then apply Marching Cubes~\cite{lorensen1998marching} to obtain the final mesh (and, optionally, $f_\text{colour}$ for the texture). 
The obtained colour can be transferred to the mesh using linear interpolation of the voxel grid colours, weighted by the SDF magnitudes; see \cref{fig:colored_mesh} in the supplementary section.

\paragraph{SDF Network} The geometry MLP $f_{\text{sdf}}$ consists of eight hidden layers (256 channels each) with Softplus activations ($\beta=100$), creating a smooth gradient landscape that stabilises SDF learning. 
We inject a skip connection between the input coordinates and the fourth layer's output to enhance high-frequency detail preservation. 
$f_{\text{sdf}}$ outputs both the SDF $\varphi(\mathbf{x}) \in \mathbb{R}$ and a 256-dimensional geometric feature vector $\mathbf{f} \in \mathbb{R}^{256}$ that encodes local shape properties. 

\paragraph{Radiance Network} For view-dependent colour prediction, we employ a shallower four-layer MLP (256 channels per layer, ReLU activations) that combines four key inputs: positionally encoded 3D coordinates ($\mathbf{x}$ with eight frequency bands), surface normals $\mathbf{n} = \nabla\varphi(\mathbf{x})$, geometric features $\mathbf{f}$ from the SDF network, and view directions $\mathbf{d}$ encoded via SHs (see \cref{sec:sh_encoding}). 
This asymmetric architecture, the deep SDF MLP for high-capacity geometric modelling paired with a shallow radiance MLP, follows the NeuS design \cite{neus} but replaces their positional view encoding with our SH-based approach to better handle event data sparsity. 
\paragraph{Implementation Details}
Our code is based on EventNeRF \cite{eventnerf}. We train each model for 600k iterations, which takes 45 GPU-hours on a single NVIDIA A100 GPU. At inference time, mesh extraction with Marching Cubes~\cite{lorensen1998marching} at $300{\times} 300{\times}300$ resolution takes ${\approx}69.3$ seconds (54.7s network queries, 8.1s surface reconstruction). 
\begin{table*}[!ht]
    \centering
    \renewcommand{\arraystretch}{1.3}
    \scalebox{0.7}{
    \begin{tabular}{l|cc|cc|cc|cc|cc|cc}
        \toprule
                & \multicolumn{2}{c|}{\textbf{Chair}} 
                & \multicolumn{2}{c|}{\textbf{Mic}} 
                & \multicolumn{2}{c|}{\textbf{Hotdog}} 
                & \multicolumn{2}{c|}{\textbf{Drums}} 
                & \multicolumn{2}{c|}{\textbf{Lego}} 
                & \multicolumn{2}{c}{\textbf{Avg}.} \\
        \midrule
        Method      & Chamfer $\downarrow$ & MAE $\downarrow$  
                    & Chamfer $\downarrow$ & MAE $\downarrow$ 
                    & Chamfer $\downarrow$ & MAE $\downarrow$ 
                    & Chamfer $\downarrow$ & MAE $\downarrow$ 
                    & Chamfer $\downarrow$ & MAE $\downarrow$
                    & Chamfer $\downarrow$ & MAE $\downarrow$\\
        \midrule
        E2VID~\cite{e2vid} + NeuS~\cite{neus} 
                    & 0.209 & 0.125   
                    & \underline{0.113} &  \underline{0.057}  
                    & 0.428 & 0.199   
                    & 0.061 & 0.034   
                    & 0.107 &  0.052 
                    & 0.184 &  0.093  \\
        EventNeRF~\cite{eventnerf}           
                    & 0.061 & 0.030   
                    & 0.298 & 0.138   
                    & \underline{0.102} & \underline{0.043}   
                    & \underline{0.054} & \textbf{0.032}   
                    & 0.084 & 0.040   
                    & 0.120 & 0.057   \\
        PAEv3D$^*$\cite{paev3d}
                    & \underline{0.050} & \underline{0.022}   
                    & 0.227 & 0.104   
                    & 0.114 & 0.050   
                    & 0.059 & 0.034   
                    & \underline{0.077} & \underline{0.036}   
                    & \underline{0.105} & \underline{0.050}   \\
        \textbf{EventNeuS (Ours)}             
                    & \textbf{0.040} & \textbf{0.017}   
                    & \textbf{0.107} & \textbf{0.052}   
                    & \textbf{0.084} & \textbf{0.037}   
                    & \textbf{0.053} & \underline{0.032}   
                    & \textbf{0.067} & \textbf{0.031}   
                    & \textbf{0.070} & \textbf{0.034}   \\
        \bottomrule
    \end{tabular}
    }
    \caption{Quantitative comparison using Chamfer Distance and MAE on NeRF synthetic scenes. 
    Our EventNeuS achieves state-of-the-art performance across all categories, with best scores \textbf{bolded} and second-best \underline{underlined}. 
    ``$^*$'': PAEv3D uses a 4x higher-resolution event stream than the rest ($692{\times}520$ px), as the method does not converge otherwise. 
    }
    \label{table:quant_results}
\end{table*}

%% file: sec/5_experiments.tex
\section{Experiments}
\label{sec:experiments}

\begin{figure*}
    \centering
    \includegraphics[width=0.98\linewidth]{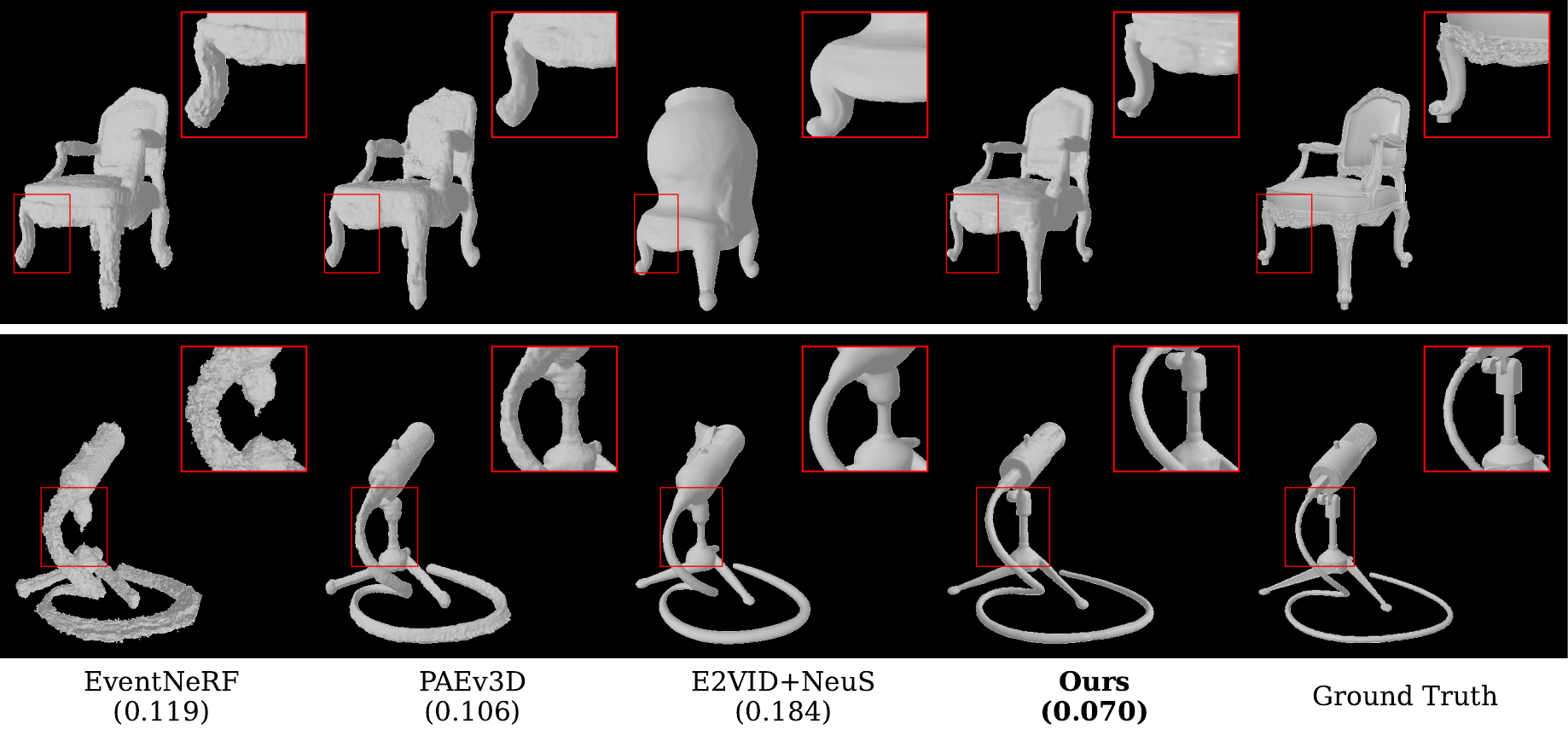}
    \vspace{-10pt}
    \caption{Qualitative comparison of our method with challenging baselines on the synthetic NeRF dataset \cite{nerf} (first row: Chair; second row: Mic). Note that all results shown are \textbf{mesh reconstructions}, not volumetric renderings. 
    }
    \label{tab:synthetic_comp}
\end{figure*}

\subsection{Datasets}
We evaluate the performance of our method on both synthetic and real-world data to validate its robustness under controlled and challenging real-world scenarios.

\subsubsection{Synthetic Dataset}\label{ssec:synthetic_data} 
We generate synthetic event streams of the synthetic Chair, Mic, Hotdog, Drums, and Lego sequences~\cite{nerf}. 
We render multi-view RGB frames at a resolution of $346{\times}260$ pixels using Blender \cite{Blender2018}, capturing each object along a spherical trajectory. 
Following a Seiffert's spherical spiral with 8 revolutions, we extract 999 frames, resulting in an average camera speed of 2.88°/frame or 2880°/s. 
Event streams are then synthesised using the event-simulator \cite{eventnerf}: 
The simulation pipeline converts sRGB images to logarithmic intensity space via gamma correction, applies an RGGB Bayer filter for colour-to-greyscale conversion, and triggers events when intensity differences exceed a fixed threshold of $C = 0.2$.
The resulting synthetic events, camera intrinsics and extrinsics are used during training (see App.~\ref{sec:syn_data}). 

\subsubsection{Real Dataset}

For real-world evaluation, we utilise the EventNeRF~\cite{eventnerf} dataset, including event streams captured with a DAVIS 346C colour event camera. 
This dataset serves as a realistic benchmark, featuring diverse structures and varying lighting conditions, to assess our model’s robustness in handling practical, real-world scenarios. 

\subsubsection{Evaluation Metrics}
\label{sec:metrics}

We use two complementary metrics: the Mean Absolute Error (MAE) of the SDF values and the Chamfer Distance (CD). 
The SDF-MAE quantifies the error between the computed SDF of the ground truth and the predicted meshes, reflecting the accuracy of the volumetric reconstruction. 
CD assesses the geometric similarity between the two meshes by measuring the average distance between points on one mesh and their nearest neighbours on the other; see App.~\ref{ssec:implementation_details} for more details. 
Jointly, SDF-MAE and CD offer a balanced evaluation: the former emphasises volumetric consistency and the accuracy of the implicit function, while the latter directly measures the closeness of the reconstructed surfaces and explicit geometric alignment. 
We further use ``MAE'' to denote SDF-MAE for compactness. 

\subsection{Comparisons and Results} 
Our evaluation highlights significant differences in the reconstruction quality between density-based and implicit surface-based approaches. 
As quantified in \cref{table:quant_results}, PAEv3D and EventNeRF exhibit noticeable jitter in the reconstructed surfaces (CD scores of 0.106 and 0.120, respectively), causing extracted meshes to appear noisy compared to our method's 0.07 average. 
This arises from insufficient constraints on 3D geometry in volume density fields (particularly in the Mic scene), where EventNeRF's CD (0.298) nearly triples our result (0.107). 
We emphasise that PAEv3D requires a high-resolution event stream $692 \times 520$ px for convergence, as indicated in the table footnotes, yet it still underperforms our approach across all categories. 
Existing implicit surface-based methods~\cite{neus}, while geometrically stable (e.g., E2VID+NeuS MAE of 0.094), rely on RGB inputs and are not directly compatible with event streams. 
Converting events to RGB frames via existing  techniques~\cite{e2vid} introduces artefacts, as shown in the Chair scene comparisons (\cref{tab:synthetic_comp}), where our method achieves 0.017 MAE versus 0.125 for the converted approach. 
EventNeuS addresses both challenges by directly optimising an implicit surface representation supervised directly using event data, as evidenced by our leading scores in 9 out of 10 cases in \cref{table:quant_results}. 
Our Eikonal loss regularises the SDF’s gradient norm, enforcing smoothness while preserving geometric consistency, which is particularly effective in the Drums scene, where we achieve 0.053 CD versus 0.054/0.059 for density-based methods. 
This dual strategy produces meshes with notably fewer artefacts---thin structures in the Chair and Mic scenes (\cref{tab:synthetic_comp}), for instance, are reconstructed with higher fidelity compared to the blurred or noisy outputs from density-based counterparts. 

\begin{table}[h]
  \centering
  \footnotesize
  \resizebox{1.0\columnwidth}{!}{
  \begin{tabular}{lccc|ccc}
    \toprule
    \multirow{2}{*}{\bfseries Scene}
      & \multicolumn{3}{c|}{\bfseries EventNeRF \cite{eventnerf}}
      & \multicolumn{3}{c}{\bfseries EventNeuS (Ours)} \\
    \cmidrule(lr){2-4} \cmidrule(lr){5-7}
      & PSNR$\!\uparrow$ 
      & SSIM$\!\uparrow$ 
      & LPIPS$\!\downarrow$ 
      & PSNR$\!\uparrow$ 
      & SSIM$\!\uparrow$ 
      & LPIPS$\!\downarrow$ \\
    \midrule
    Chair      & \textbf{33.05} & 0.98 & 0.06 & 30.94 & \textbf{0.99} & \textbf{0.04} \\
    Mic       & 29.61 & 0.98 & 0.02 & \textbf{30.57} & 0.98 & 0.02 \\
    Lego      & 23.81 & 0.95 & 0.08 & \textbf{24.34} & \textbf{0.96} & \textbf{0.06} \\
    Drums      & 27.83 & 0.97 & 0.05 & \textbf{28.65} & \textbf{0.98} & \textbf{0.03} \\
    Hotdog        & \textbf{29.40} & 0.97 & 0.08 & 28.35 & \textbf{0.98} & \textbf{0.07} \\
    \midrule
    \textbf{Average}
               & \textbf{28.74} & 0.97 & 0.06 & 28.57 & \textbf{0.98} & \textbf{0.04} \\
    \bottomrule
  \end{tabular}
  }
  \caption{\textbf{Complementary novel view synthesis evaluation on synthetic scenes.} While EventNeRF achieves marginally higher PSNR in two cases, EventNeuS demonstrates superior perceptual quality through improved SSIM and LPIPS scores.} 
  \label{tab:novel_view_metrics}
\end{table}

Table \ref{tab:novel_view_metrics} additionally reports complementary novel-view synthesis metrics. 
See App.~\ref{ssec:novel_view_sythesis} for a detailed discussion. 
Note that we were not able to include EvAC3D~\cite{evac3d} in the comparisons due to engineering reasons and incomplete source code documentation. 
\begin{figure}
    \centering
    \includegraphics[width=1.0\linewidth]{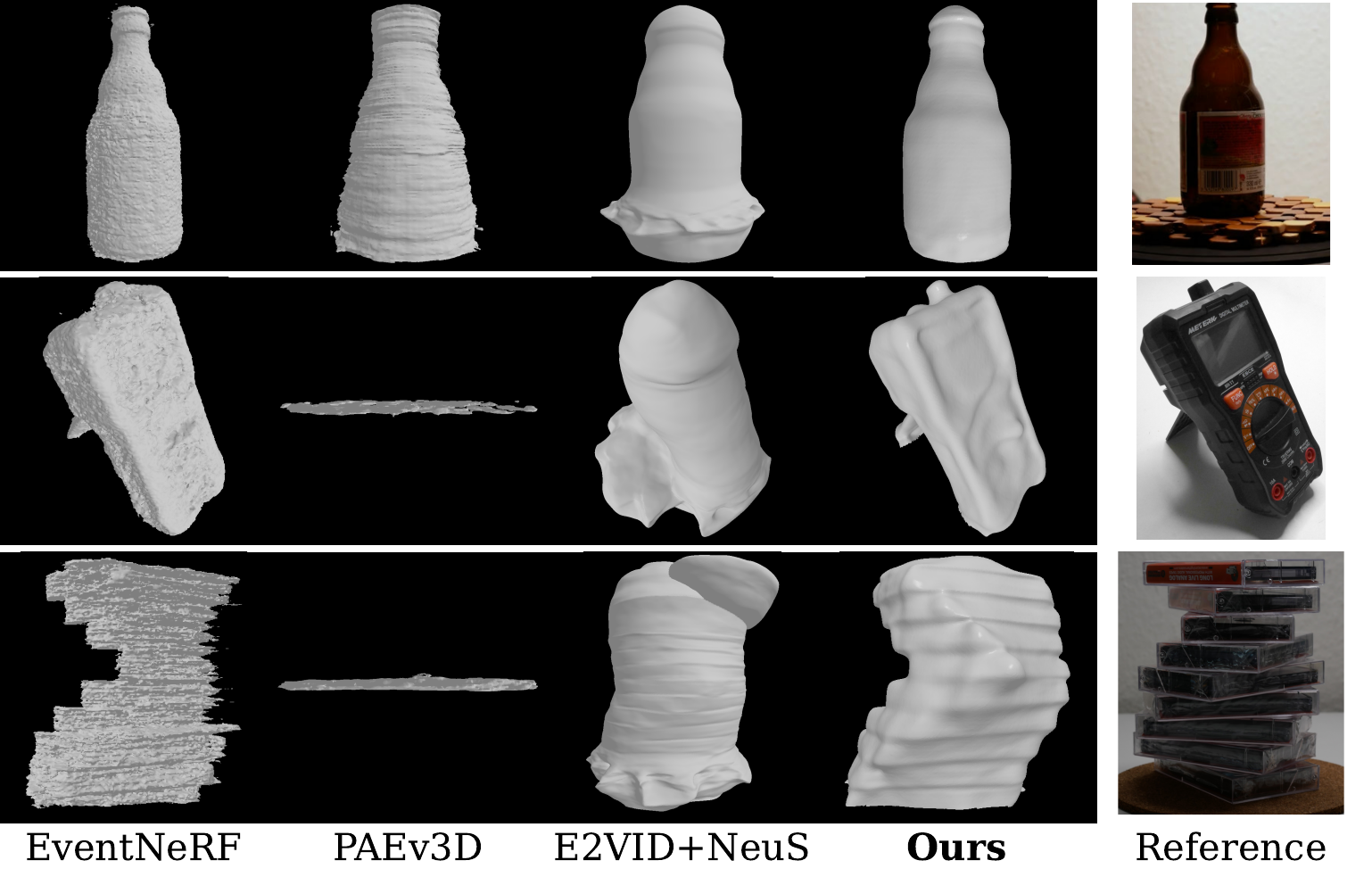}
    \caption{Qualitative comparison on the real dataset \cite{eventnerf} with fast-rotating objects observed by an event camera. 
    All results shown are \textbf{mesh reconstructions}, not volumetric renderings. The RGB views are provided for reference only. 
    Best viewed with zoom.
    } 
    \label{tab:real_comp}
\end{figure}

\cref{tab:real_comp} shows that EventNeuS accurately reconstructs thin structures (e.g., multimeter and tape ridges) that baselines blur or miss, producing smoother surfaces due to robust implicit representation and effective regularisation.
This not only enables the recovery of fine geometric details but also ensures consistency across diverse scenes. 
These advantages make EventNeuS particularly well-suited for event-based 3D reconstruction tasks, where precise surface recovery is essential.

\begin{table}[ht]
    \centering
    \scalebox{0.82}{
    \begin{tabular}{lcc}
        \toprule
        \textbf{Method} & \textbf{Chamfer Distance $\downarrow$} & \textbf{MAE $\downarrow$} \\ 
        \midrule
        Without Negative Sampling & 0.080 & 0.044 \\
        Without Eikonal Loss      & 0.077 & 0.044 \\
        Without SH Encoding       & 0.082 & 0.043 \\
        Without Annealing         & 0.079     & 0.042     \\
        \midrule
        PE Frequencies, nine steps & 0.098 & 0.053 \\
        PE Frequencies, five steps & 0.248 & 0.120 \\
        \midrule
        \textbf{Full Method} & \textbf{0.071} & \textbf{0.034} \\
        \bottomrule
    \end{tabular}}
    \caption{Ablation study and design choice evaluation using both CD and MAE. We report performance for several model variants after disabling negative sampling, Eikonal loss, SH encoding, and annealing strategies. Notably, the full method achieves the best performance with CD of 0.071 and MAE of 0.034, highlighting the importance of each component of our EventNeuS approach.} 
    \label{tab:ablation_results}
\end{table}

\begin{figure}[h]
  \centering
  \vspace{-10pt}
  \setlength{\tabcolsep}{4pt}
  \begin{tabular}{cccc}
    \includegraphics[width=0.22\linewidth]{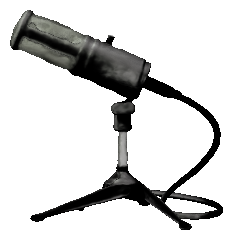} &
    \includegraphics[width=0.22\linewidth]{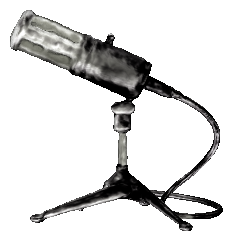} &
    \includegraphics[width=0.22\linewidth]{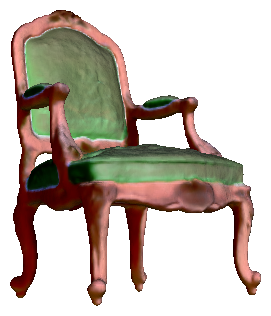} &
    \includegraphics[width=0.22\linewidth]{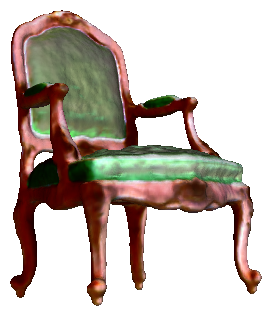} \\
    \multicolumn{2}{c}{\parbox{0.49\linewidth}{\centering (a) Mic}}
      &
    \multicolumn{2}{c}{\parbox{0.49\linewidth}{\centering (b) Chair}}
  \end{tabular}
  \vspace{-5pt}
  \caption{SH encoding effectiveness on textured meshes (left: with SH, right: without SH). SH enable more detailed surfaces and better representation of view-dependent effects.}
  \label{fig:sh_ablation}
\end{figure}

\subsection{Ablation Study} 
We further assess the contribution of key components in EventNeuS through an ablation study, as summarised in Table~\ref{tab:ablation_results}. Removing negative sampling increases the CD and the MAE. This indicates that negative sampling is crucial for suppressing false surface detections by effectively modelling free space. Similarly, excluding the Eikonal loss leads to a slight degradation, demonstrating its importance in enforcing a smooth and well-behaved implicit surface.

The study further reveals that SH encoding is vital for capturing high-frequency appearance variations and fine geometric details. Without it (using standard positional encoding for view directions instead), the metrics deteriorate, suggesting that a more expressive view-dependent representation is crucial.
Fig.~\ref{fig:sh_ablation} provides a qualitative comparison showing the effectiveness of SH encoding in the rendered textured meshes, where scenes with SH encoding exhibit superior surface detail and appearance quality. 
Moreover, the positional encoding (PE) frequency plays a significant role in refining the surface convergence. Experiments with different PE frequencies show that a higher frequency (nine steps) produces better accuracy compared to fewer steps. 
This analysis highlights the synergetic effect of our design choices and their critical role in achieving high-fidelity, event-based 3D reconstruction.

%% file: sec/6_conclusions.tex
\section{Discussion and Conclusion}
\label{sec:conclusion}

\begin{figure}
    \centering
    \includegraphics[width=0.75\linewidth,trim={0cm 0cm 0cm 0cm},clip]{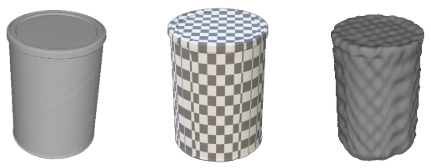}
    \vspace{-5pt}
    \caption {The left image is the ground-truth mesh, and the middle image shows the ground-truth mesh after applying chequerboard textures. The right image shows the mesh recovered with our method, with noticeable texture misinterpretation artefacts.} 
    \label{fig:limitation} 
      
\end{figure}

As we are accumulating the events over a temporal window, we are missing some of the high-frequency temporal information in complex textures and lighting. While random window sampling solves this problem to some extent, there is still room for further research. However, incorporating more refined temporal-spatial adaptation strategies could further enhance reconstruction quality, especially in regions with high-frequency changes. We also encounter artefacts on the reconstructed mesh, such as unintended texture imprints, due to the nature of the implicit surface reconstruction method (see \cref{fig:limitation}). This issue arises because the implicit surface model learns the neural surface based on texture features. Additionally, as event cameras generate events based on changes in the brightness intensity of the texture, these artefacts persist in our results as well. Our method currently does not support reconstructing large-scale scenes due to fundamental constraints, including limited network capacity for modelling expansive geometries and increased uncertainty in camera pose estimation.

\noindent \textbf{Conclusion.} 
In summary, we show that it is possible to reconstruct a 3D mesh of the observed rigid object with high accuracy, relying on the volumetric formulation solely from a monocular (possibly fast-moving) event camera. 
Our method achieves lower CD and SDF-MAE metrics across various objects compared to previous state-of-the-art methods, indicating closer alignment between the reconstructed and ground-truth surfaces. 
Qualitative assessments further confirm EventNeuS's ability to capture fine geometric details and produce smooth, accurate surface reconstructions. 
Future work can explore the integration of recent advances in RGB-based neural surface estimation---combined with 3D Gaussian Splatting (3DGS) \cite{Guedon2024}---into event-driven differentiable rendering frameworks, as 3DGS has been shown to be compatible with event-based supervision \cite{nakabayashi2025ev4dgs}.

%% file: sec/X_supp.tex
\clearpage
\appendix
\maketitlesupplementary

\begin{figure}
    \centering
    \includegraphics[width=0.8\linewidth,trim={0cm 0cm 0cm 0cm},clip]{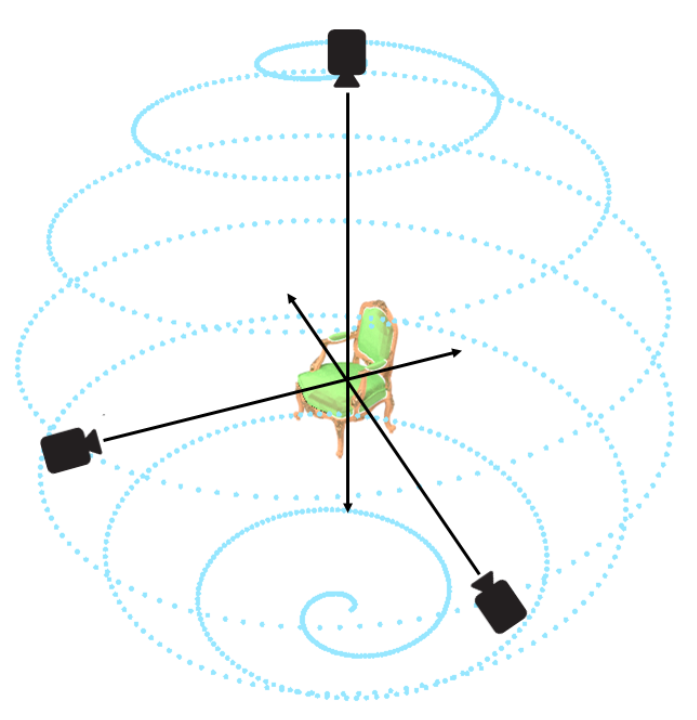}
    \caption {Visualisation of Seiffert's spherical spiral trajectory of the virtual monocular event camera used in our synthetically generated data. See Sec.~\ref{ssec:synthetic_data} and App.~\ref{sec:syn_data} for further details. 
    }
    \label{fig:trajectory}
\end{figure}

This supplementary document provides additional details on the datasets (App.\ref{sec:syn_data}), implementation details (App.~\ref{ssec:implementation_details}), and more visualisations of the qualitative experimental results (App.~\ref{sec:results}). 

\section{Synthetic Data Generation}
\label{sec:syn_data}

We generate synthetic event streams using ground truth meshes from the NeRF~\cite{nerf} synthetic datasets. The 3D assets are imported into Blender's Cycles renderer, where we render 999 uniformly sampled views along Seiffert's spherical spiral trajectory, \cref{fig:trajectory}. Our capture sequence comprises eight complete revolutions around each object, beginning at the base orientation and ascending to the apex. Each RGB frame is rendered at $346\times260$ pixels resolution with HDR lighting to simulate realistic intensity variations.

Event streams are synthesised using ESIM~\cite{eventnerf}, an event camera simulator that computes per-pixel logarithmic brightness changes. We configure ESIM with the following parameters: 
\begin{verbatim}
    "Resolution": "260, 346"
    "bg_val": "159.0/255.0"
    "THR": "0.2"
    "gamma": "2.2"
\end{verbatim}

\begin{figure*}
    \centering
    \includegraphics[width=1.0\linewidth]{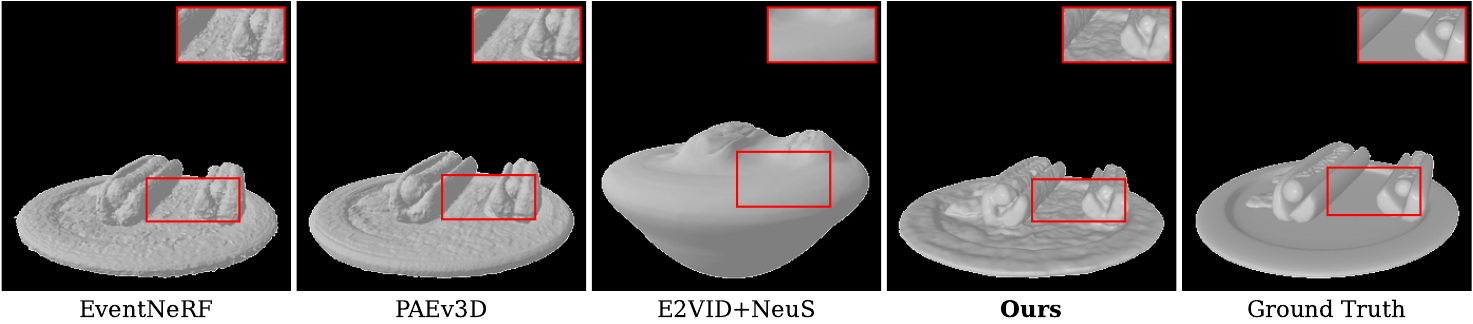}
    \caption {Qualitative comparison on the Hotdog scene. E2VID+NeuS~\cite{e2vid,neus} produces oversmoothed geometry that loses fine structural details. The results of  EventNeRF~\cite{eventnerf} and PAEv3D~\cite{paev3d} both suffer from noticeable surface jitter artefacts, particularly evident along the sausage curvature, where the density-based representations fail to recover smooth surface. Our EventNeuS produces a significantly cleaner reconstruction, accurately preserving the surface smoothness and characteristic shape of the hotdog with fewer artefacts.} 
    \label{fig:results_hotdog}
\end{figure*}

\begin{figure*}
  \centering
  \setlength{\tabcolsep}{2pt}
  \begin{tabular}{cccc}
    \includegraphics[width=0.21\linewidth]{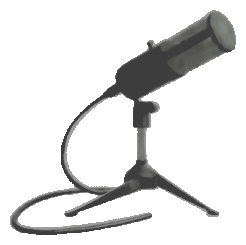} &
    \includegraphics[width=0.21\linewidth]{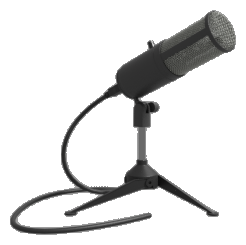} &
    \includegraphics[width=0.27\linewidth]{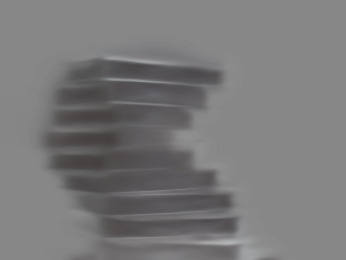} &
    \includegraphics[height=0.205\linewidth,keepaspectratio=true]{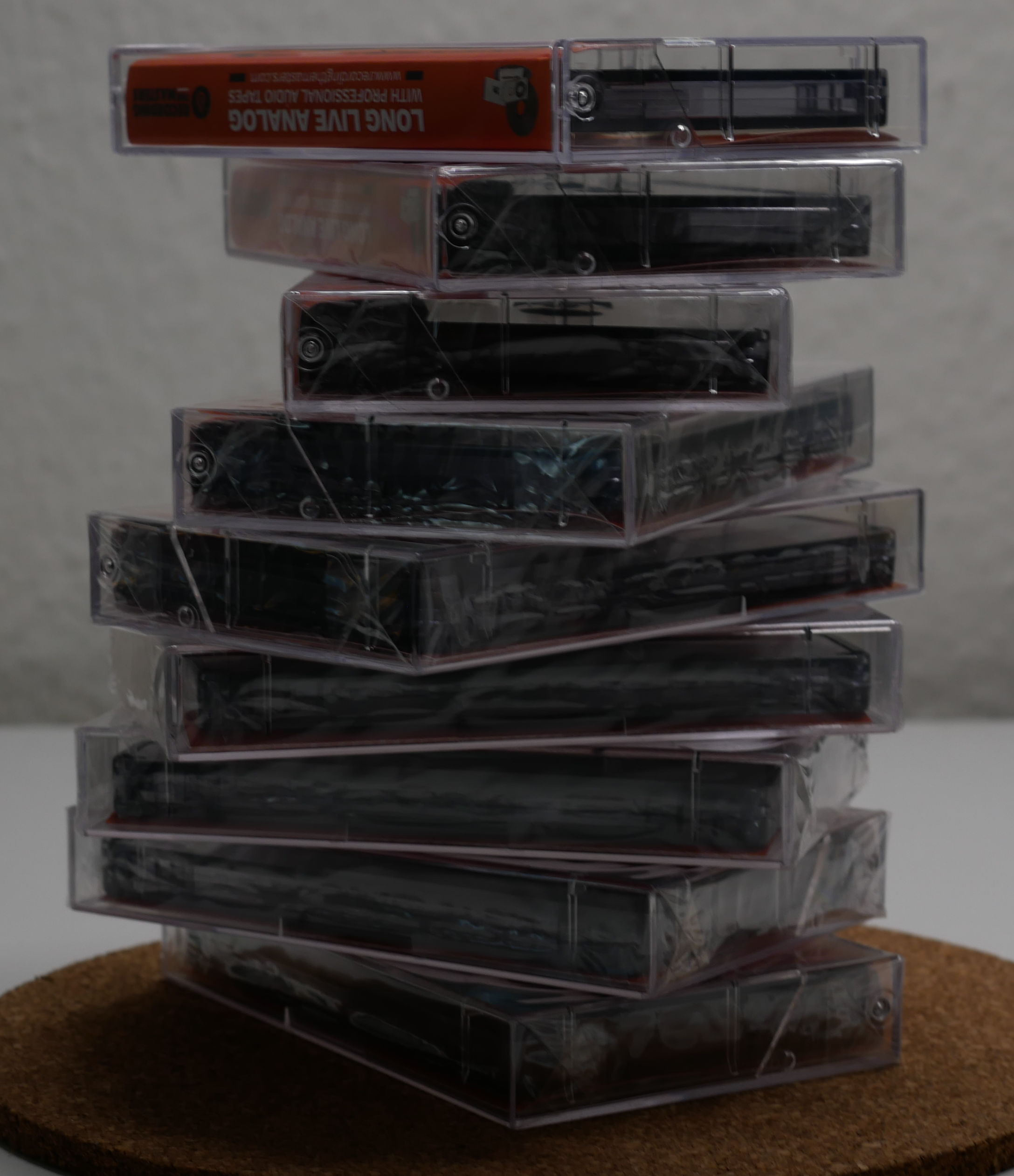} \\[-4pt]
    \small Mic (Ours) & \small Mic (Reference) & \small Tape (Ours) & \small Tape (Reference) \\[10pt]
    \includegraphics[width=0.21\linewidth]{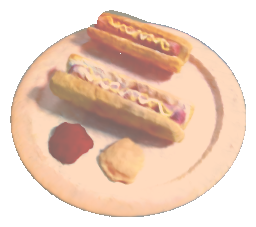} &
    \includegraphics[width=0.21\linewidth]{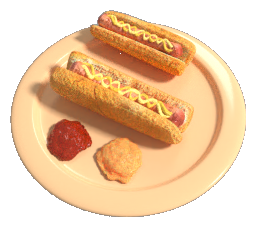} &
    \includegraphics[width=0.27\linewidth]{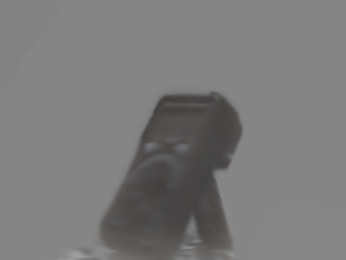} &
    \includegraphics[width=0.27\linewidth]{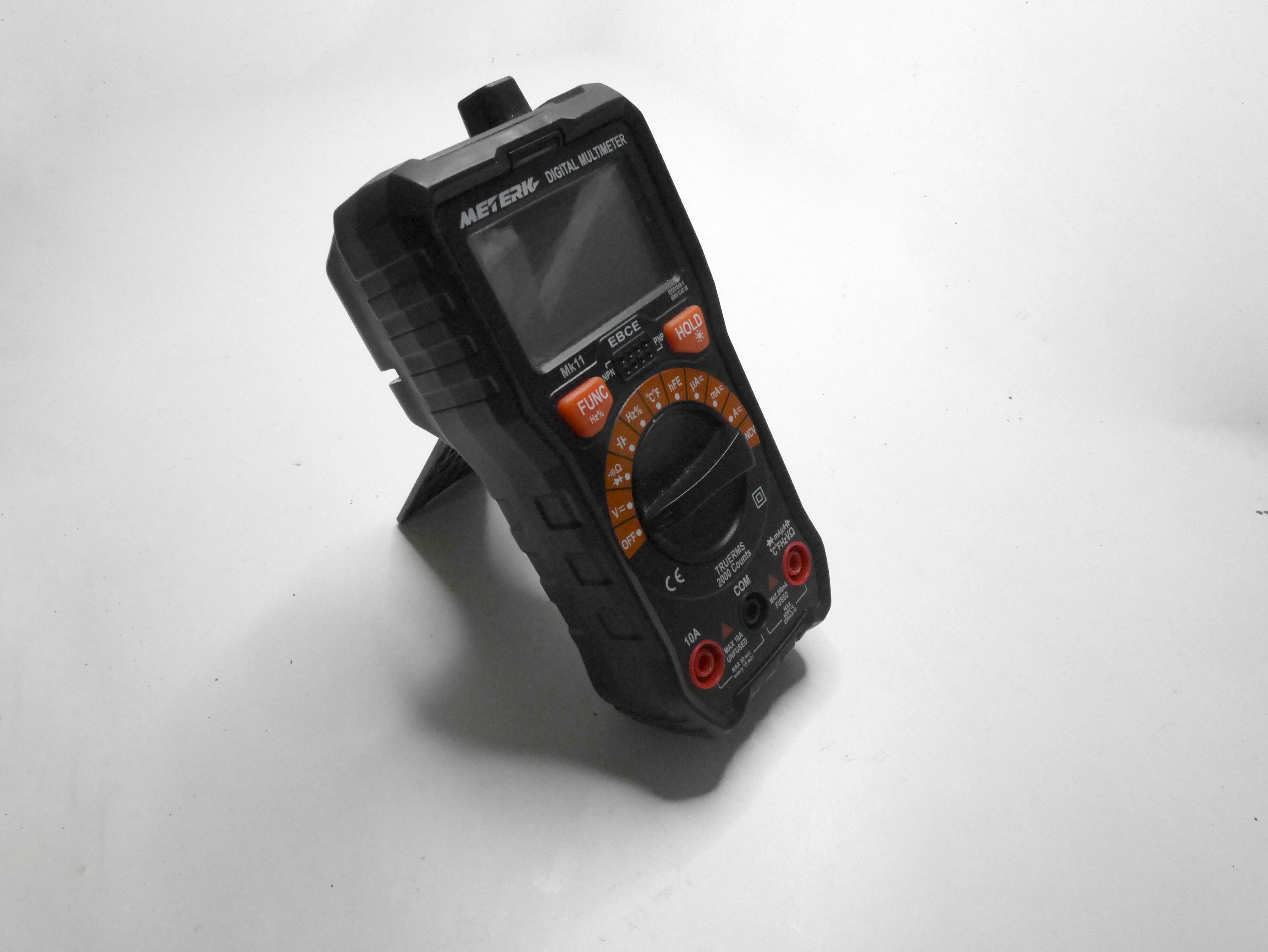} \\[-4pt]
    \small Hotdog (Ours) & \small Hotdog (Reference) & \small Multimeter (Ours) & \small Multimeter (Reference) \\
  \end{tabular}
  \vspace{-5pt} 
  \caption{Novel-view synthesis results demonstrating EventNeuS's capability to render high-quality views from learned geometry, showcasing both synthetic and real-world scenarios. Reference RGB images are shown for reference only and are not used during training.}
  \label{fig:novel_views}
\end{figure*}

\section{Implementation Details}
\label{ssec:implementation_details}

\noindent \textbf{SH Encoding.} 
To model view-dependent colour variations efficiently, we encode camera directions $\mathbf{d}$ using TinyCudaNN's ~\cite{tiny-cuda-nn} optimised SH implementation. This produces a compact 16D feature vector.
For a direction \(\mathbf{d}\), we compute:
\[
\text{SH}(\mathbf{d}) = \texttt{tcnn.SphericalHarmonics}(\mathbf{d}) \in \mathbb{R}^{16}
\]

For camera directions $\mathbf{d}$, we configure the encoder with:

\begin{verbatim}
encoding_config = {
    "otype": "SphericalHarmonics",  
    "degree": 4,                    
}
\end{verbatim}

These 16D SH bases are further concatenated with surface normals $\nabla\varphi(\mathbf{x})$ and geometric features from the SDF network $\varphi(\mathbf{x})$ to predict view-dependent colours.
\vspace{5pt} 
\noindent \textbf{Evaluation Metrics.} 
To compute the SDF-MAE, we implement a hybrid sampling strategy: 50\% of the evaluation points are sampled directly from the mesh surfaces with slight random perturbations (within $\pm$1\% of the mesh’s maximum dimension) to capture fine surface details, while the remaining 50\% are uniformly sampled within a bounding box that encloses both meshes. For each sampled point, we calculate the absolute difference between the ground truth and predicted SDF values.

To compute CD, both meshes are first normalised to a unit cube to ensure scale invariance. We then uniformly sample $10^4$ points from each mesh’s surface and perform nearest-neighbour search using a KD-Tree. This bidirectional metric penalises both missing and extraneous geometry, effectively capturing discrepancies in surface alignment. 

\section{Additional Visualisations}
\label{sec:results}

We provide comprehensive evaluations across standard NeRF synthetic scenes. As quantitatively demonstrated in Table~\ref{table:quant_results}, EventNeuS achieves superior reconstruction quality, particularly evident in geometrically complex structures like the Hotdog's sausage, where our method preserves thin protrusions and surface curvature with 0.0839 Chamfer distance versus 0.1024-0.4281 for baselines. Qualitative comparisons (\cref{fig:results_hotdog}) further reveal that EventNeuS preserves more surface details than any other baseline.

\subsection{Novel-View Synthesis Evaluation}\label{ssec:novel_view_sythesis}
While our work primarily focuses on geometry reconstruction, we additionally evaluate the novel view synthesis capabilities of EventNeuS. We compute PSNR, SSIM, and LPIPS metrics on regions of interest (ROIs) using ground-truth foreground masks for both EventNeuS and EventNeRF. As shown in \cref{tab:novel_view_metrics} (main paper), EventNeuS achieves competitive performance across all metrics, with particularly notable improvements in SSIM and LPIPS. An important conclusion from these results is that better geometry facilitates more accurate novel views, as evidenced by our consistent improvements in perceptual quality metrics. Fig.~\ref{fig:novel_views} shows qualitative examples of our novel view synthesis results on both synthetic and real data, demonstrating EventNeuS's capability to render high-quality views from the learned geometry.

\begin{figure}[ht]
  \centering
  \begin{subfigure}[t]{0.41\linewidth}
    \centering
    \includegraphics[width=\linewidth]{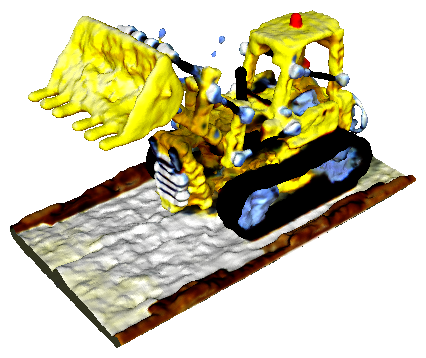}
  \end{subfigure}
  \begin{subfigure}[t]{0.41\linewidth}
    \centering
    \includegraphics[width=\linewidth]{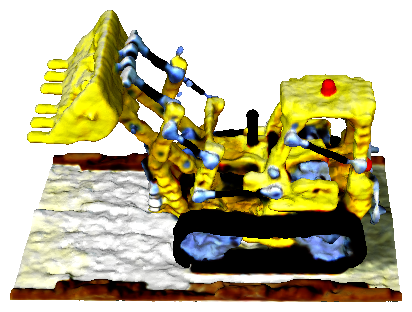}
    \label{fig:image2}
  \end{subfigure}
  \caption{Textured mesh of the Lego scene obtained using our EventNeuS approach, shown from two arbitrary viewpoints.} 
  \label{fig:colored_mesh}
\end{figure}

\subsection{Coloured Meshes} 
Once EventNeuS is trained, we extract the final mesh by querying both the SDF network \(f_{\text{sdf}}\) and the colour field \(f_{\text{colour}}\) on a dense grid.
We then apply the Marching Cubes algorithm~\cite{lorensen1998marching} to generate vertices from the SDF grid, while simultaneously interpolating vertex colours from the colour grid.
The interpolation is weighted by the inverse SDF values to prioritise points closer to the surface.
This process yields a high-fidelity textured mesh, as demonstrated for the Lego scene in \cref{fig:colored_mesh}.
The rendered views exhibit sharp geometric features and consistent colouration, such as the crisp edges and uniform surfaces of the Lego bricks.
This result underscores our method's capability to jointly reconstruct precise geometry and detailed appearance solely from an event stream.